\pdfoutput=1

\documentclass[11pt]{article}

\usepackage{style/EMNLP2023}

\usepackage{times}
\usepackage{latexsym}
\usepackage{booktabs}

\usepackage[T1]{fontenc}

\usepackage[utf8]{inputenc}

\usepackage{microtype}

\usepackage{inconsolata}

\usepackage{listings} 
\usepackage{graphicx}
\usepackage{amssymb} 
\usepackage{mathtools} 
\usepackage{pgfplots}
\usepackage{tikz}
\usepackage[inkscapelatex=false]{svg}
\usepackage{amsmath}

\usetikzlibrary{external}
\pgfplotsset{compat=1.18}

\title{Data-Efficient Domain Adaptation for LLM-based MT using Contrastive Preference Optimization}

\author{
Inacio Vieira\\
Alpha CRC, Cambridge, UK\\
\texttt{ivieira@alphacrc.com}
\And
Antonio Castaldo\\
University of Naples “L’Orientale”\\
University of Pisa\\
\texttt{antonio.castaldo@phd.unipi.it}
\AND
James O’Doherty\\
Independent Researcher\\
\texttt{jamesjodoherty@gmail.com}
\And
Sheila Castilho\\
SALIS / ADAPT Centre\\
Dublin City University, Ireland\\
\texttt{sheila.castilho@adaptcentre.ie}
}

\begin{document}
\maketitle
\begin{abstract}
LLMs often require adaptation to domain-specific requirements, a process that can be expensive when relying solely on SFT. We present an empirical study on applying CPO to simulate a post-editing workflow for data-efficient domain adaptation. Our approach synthesizes preference pairs by treating the base model's own raw output as the 'rejected' translation and the human-approved TM entry as the 'chosen' one. This method provides direct feedback on the model's current knowledge, guiding it to align with domain-specific standards. Experiments in English–Brazilian Portuguese and English–Korean show that, by using just 14.7k preference pairs, the model achieves performance close to that of a model trained on 160k+ samples with SFT, demonstrating significant data efficiency. Although we showcase its effectiveness in MT, this application of CPO naturally generalizes to other generative tasks where a model’s initial drafts can serve as a contrastive signal against a golden reference.
\end{abstract}

\section{Introduction}

Large Language Models (LLMs) have reshaped many Natural Language Generation (NLG) tasks, including Machine Translation (MT). They show impressive performance in general domain content \citep{alves-etal-2023-steering, hendy_how_2023, robinson-etal-2023-chatgpt, zhu_multilingual_2023, lyu_paradigm_2024}, but often falter in specialized domains or when tasked with meeting strict style and terminology demands \citep{zheng_fine-tuning_2024}. 
In these scenarios, \emph{domain adaptation} becomes vital. Organizations commonly leverage proprietary data, such as translation memories (TMs), to tailor smaller, business-friendly LLMs \citep{touvron_llama_2023, llama_team_llama_2024} without incurring the high costs of large-scale retraining \citep{moslem_adaptive_2023}.

However, questions remain regarding how best to exploit TMs for fine-tuning. While Supervised Fine-Tuning (SFT) is a common approach, recent work on preference optimization suggests that contrastive methods can offer a more potent signal for alignment \citep{rafailov_direct_2024, xu_contrastive_2024}. These methods train a model to distinguish between "chosen" (preferred) and "rejected" (dispreferred) outputs.

Inspired by this, we investigate a specific application of Contrastive Preference Optimization (CPO) for MT domain adaptation. Our motivation draws from constructivist learning theory \citep{piaget1952origins}, where a learner's prior knowledge is elicited and then refined through corrective feedback. Analogously, we treat the LLM's raw, out-of-the-box translation as its "prior knowledge". We then create a preference pair by contrasting this raw output (the `rejected' candidate) with a high-quality, human-approved TM entry (the `chosen' candidate). This process simulates a post-editing workflow, where the goal is to close the gap between the initial machine translation and the desired final output.

Our approach differs from other preference-based methods in a few key ways. Unlike work that learns a separate reward model with RLHF \citep{xu2024advancing}, we use CPO for direct, single-stage fine-tuning, which is simpler and more computationally efficient. Furthermore, our method generates preference pairs using the model's current, on-policy outputs, providing immediate feedback on its performance. This contrasts with approaches that rely on historical, offline logs of post-edits \citep{berger2025posteditspreferences}, and it opens the door for more dynamic, iterative training cycles.

Our main contributions are empirical:
\begin{itemize}
    \item We demonstrate that applying CPO with synthetically generated pre-edits is a highly data-efficient method for MT domain adaptation, achieving with 14.7k examples what SFT requires over 160k examples for.
    \item We provide a direct, controlled comparison between this CPO-based approach and a strong SFT baseline on two language pairs, quantifying the performance gains.
    \item We show that this setup, which treats human-approved TMs as the definitive "chosen" reference, is effective for aligning an LLM with specific terminological and stylistic requirements.
\end{itemize}

For direct comparability with prior work in MT domain adaptation (Anonymous, 2024), we use identical dataset splits and subsets to compare our CPO-based approach against SFT on \texttt{Llama-3-8B-Instruct} \citep{dubey_llama_2024}. Our experiments on English–Korean and English–Brazilian Portuguese show that our method achieves comparable performance to a 160k+ SFT model with only 14.7k examples. For brevity, we summarize the full SFT training sets as "160k+"; exact sizes are 217{,}555 (PT-BR) and 162{,}360 (KO). In short, creating preference pairs from the model’s own outputs against existing TMs can substantially reduce the amount of data required for domain adaptation.

\begin{figure*}[h]
    \centering
    \includegraphics[width=\linewidth]{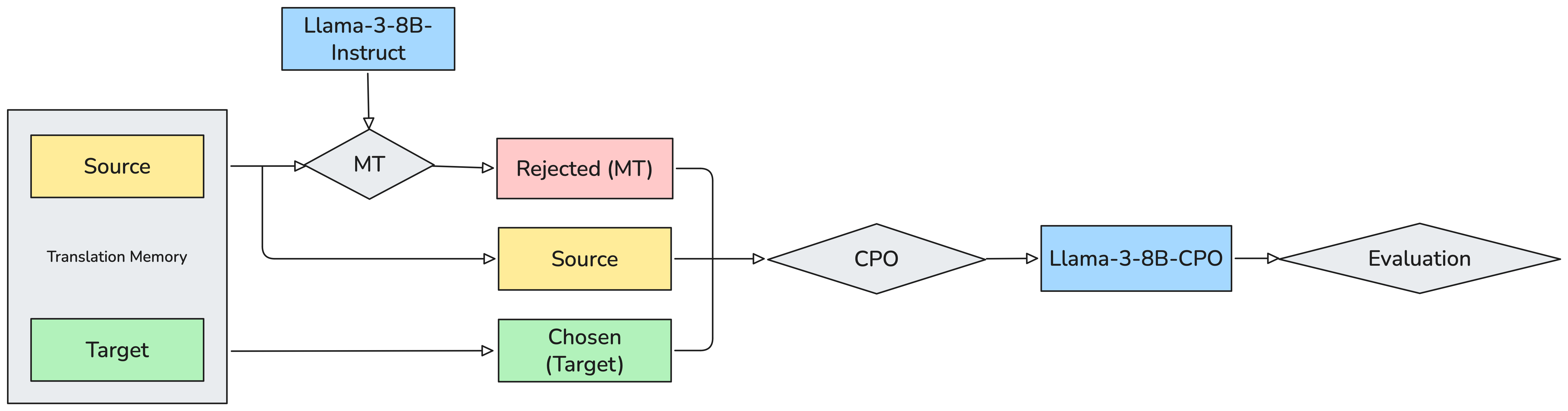}
    \caption{End-to-end workflow showing (1) synthetic preference pair generation through base model inference, (2) CPO fine-tuning, and (3) automatic evaluation.}
    \label{fig:cpo}
\end{figure*}

\section{Related Work}

Domain adaptation has long been crucial for MT systems \citep{chu2018surveydomainadaptationneural, sennrich-etal-2017-university}. Post-editing (PE) workflows, where human translators correct raw MT outputs, are standard practice for ensuring quality \citep{tatsumi-2009-correlation, moorkens_translators_2018, gladkoff-han-2022-hope}. Our work aims to leverage the insights from this human-in-the-loop process to improve the underlying MT model more efficiently.

Recent years have seen a surge in preference-based fine-tuning methods for LLMs. Reinforcement Learning from Human Feedback (RLHF) learns a reward model from human preferences and then optimizes the LLM using RL \citep{christiano2017deep}. Direct Preference Optimization (DPO) simplifies this by directly optimizing the LLM to satisfy preferences without an explicit reward model \citep{rafailov_direct_2024}. Our work builds on CPO from \citet{xu_contrastive_2024}, which extends these ideas.

Several recent works have applied preference optimization to MT. \citet{xu2024advancing} use an RLHF-style approach, training a reward model on pairs of human and machine translations. In contrast, our method uses CPO for direct optimization, avoiding the need for a separate reward model and offering a simpler, single-stage training pipeline.

Our work is closely related to \citet{berger2025posteditspreferences}, who also use CPO on post-edited data. However, a key distinction lies in the source of the preference data. They use historical, offline logs of MT outputs and their corresponding human post-edits. Our approach, instead, generates the 'rejected' candidate using the *current* model's output (i.e., on-policy). This provides direct feedback on the model's present state, aligning with constructivist principles and enabling iterative training paradigms where the model can be continuously improved with fresh outputs.

Finally, our application of CPO differs slightly from the original formulation by \citet{xu_contrastive_2024}. Their method can use automatic metrics to decide if a model-generated output is superior to a human reference, making the preference conditional. We, however, operate under the assumption that for domain adaptation, the vetted, human-approved TM is always the "chosen" gold-standard reference. This grounds the model firmly in the specific terminology and style required by the domain.

\section{Methodology}

\begin{figure}[h]
    \centering
    \includegraphics[width=\linewidth]{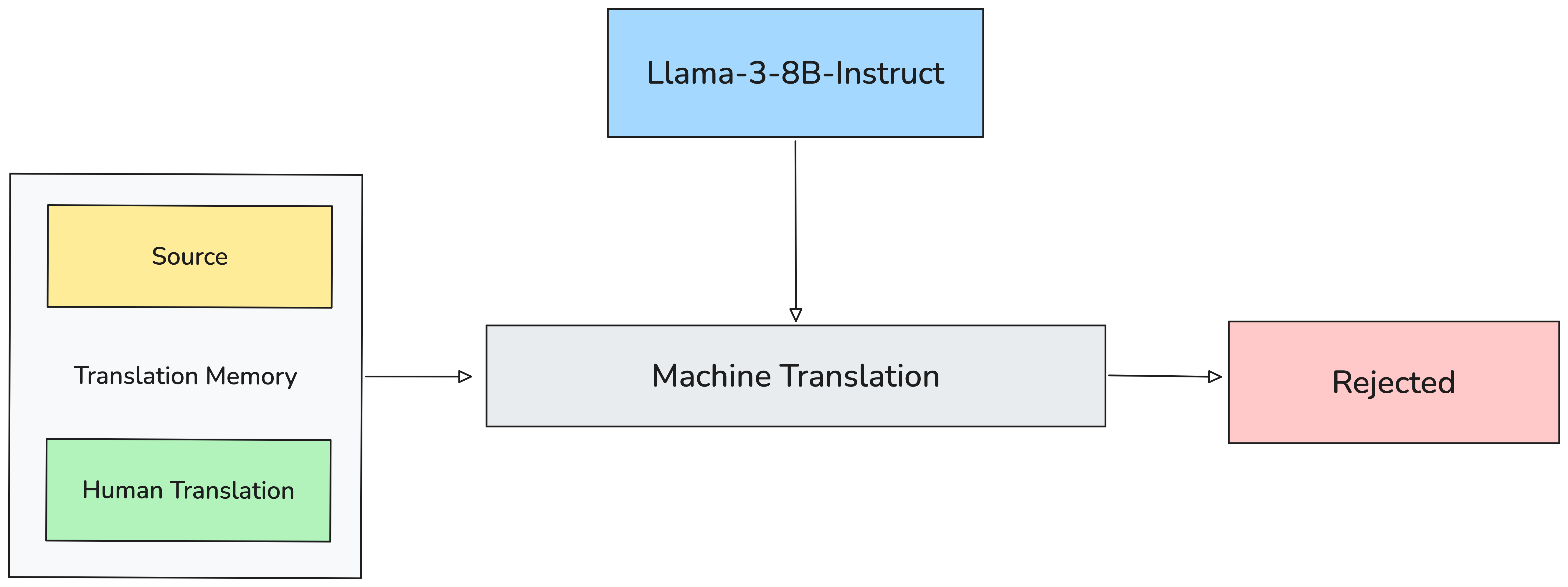}
    \caption{Generating the synthetic 'rejected' candidate by running inference with TM source text on the base model.}
    \label{fig:synthetic}
\end{figure}

We propose a contrastive training approach that simulates a PE workflow using triplets of \emph{source text, synthetic 'rejected' translation, and human-approved 'chosen' translation}. The method contrasts the translation produced by the raw base model with the final, domain-aligned reference from a TM. Figure~\ref{fig:cpo} summarizes the full pipeline, and Figure~\ref{fig:synthetic} illustrates how the synthetic 'rejected' candidate is generated.

\subsection{Simulating Post-Edits for Preference Pairs}
\label{subsec:pemt-workflow}

In a standard PE workflow, a human translator refines a raw system output. We simulate this process to generate our training data:
\begin{enumerate}
    \item For each source segment $x$ from the TM, we run inference on the baseline LLM to get its out-of-the-box translation, which we label $y_{\text{rejected}}$.
    \item We pair $y_{\text{rejected}}$ with $y_{\text{chosen}}$, the trusted TM entry for $x$.
    \item The CPO objective then trains the model to assign a higher probability to $y_{\text{chosen}}$ than to $y_{\text{rejected}}$, effectively teaching the model to prefer the domain-specific translation over its initial, unaligned output.
\end{enumerate}

By treating the LLM’s own "prior" translation as the rejected candidate, we explicitly model the correction process. This setup does not depend on any additional automatic metrics; we assume the domain-vetted TMs are intrinsically valid references that embody the required style and terminology.

\paragraph{Example Workflow}
Consider the following simplified example from our English–Brazilian Portuguese TM dataset:

\begin{itemize}
    \item \textbf{Source:} \textit{"Press the Save button to store changes."}
    \item \textbf{Human-Approved TM Entry (`chosen'):} \textit{"Pressione o botão Salvar para armazenar as alterações."}
    \item \textbf{Base Model Output (`rejected'):} \textit{"Pressione o botão Salvar para gravar mudanças."}
\end{itemize}

Our method captures the correction required here by training the model to prefer \textit{"armazenar as alterações"} over \textit{"gravar mudanças"}. Each training instance thus resembles a mini post-edit scenario where $y_{\text{rejected}}$ is contrasted with $y_{\text{chosen}}$.

\subsubsection{Preference Data Construction}
\label{sec:pref-data}

Formally, for each source sentence $x_i$, we generate two candidate translations,
\[
  Y_i \;=\; \bigl\{y_i^{\text{rejected}}, \; y_i^{\text{chosen}}\bigr\}.
\]
Here, $y_i^{\text{rejected}}$ is the LLM’s raw greedy output, and $y_i^{\text{chosen}}$ is a high-quality translation from our domain-specific TM. We then define a preference dataset
\[
  D_{\text{pref}}
    \;=\;\Bigl\{\bigl(x_i,\;y_i^{r},\;y_i^{c}\bigr)\Bigr\}_{i=1}^{N},
\]
where
\[
  y_i^{r} \;=\; y_i^{\text{rejected}}, 
  \quad
  y_i^{c} \;=\; y_i^{\text{chosen}}.
\]
Thus, $\,(y_i^{r},\,y_i^{c})$ forms a rejected/chosen pair for each $x_i$.

\subsection{Implementation Overview}

We train our model using a single-stage CPO setup, where each data point is the triplet $(x, y_{\text{chosen}}, y_{\text{rejected}})$ described above. We follow a procedure similar to prior work (Anonymous, 2024) to generate $y_{\text{rejected}}$, ensuring consistent prompts and post-processing. Afterwards, we fine-tune the \texttt{Llama-3-8B-Instruct} checkpoint using LoRA-based low-rank adaptation \citep{hu_lora_2021} and 4-bit quantization \citep{dettmers_qlora_2023}.

\paragraph{Training Objective:}
The CPO loss encourages the model to rank $y_{\text{chosen}}$ above $y_{\text{rejected}}$ while simultaneously maximizing the likelihood of $y_{\text{chosen}}$ under the model’s distribution. Concretely, the objective is:
\[
  L_{\mathrm{CPO}} \;=\; L_{\mathrm{pref}} \;+\; L_{\mathrm{SFT}},
\]
where $L_{\mathrm{pref}}$ is a preference-ranking term, and $L_{\mathrm{SFT}}$ is the negative log-likelihood on $y_{\text{chosen}}$. 

\paragraph{Preference Loss:}
Formally, let $\pi_{\theta}$ be our model’s conditional distribution. For each triplet 
$(x_i, y_i^{r}, y_i^{c}) \in D_{\text{pref}}$, we define:

\begin{align}
L_{\mathrm{pref}} &=
-\frac{1}{N} \sum_{i=1}^{N}
\Bigl[
  \log \sigma\Bigl(
    \beta \,\log 
    \frac{\pi_{\theta}\bigl(y_i^{c}\mid x_i\bigr)}
         {\pi_{\theta}\bigl(y_i^{r}\mid x_i\bigr)}
  \Bigr)
\Bigr],
\label{eq:pref} \\[4pt]
L_{\mathrm{SFT}} &=
-\frac{1}{N} \sum_{i=1}^{N}
\bigl[\log \,\pi_{\theta}\bigl(y_i^{c}\mid x_i\bigr)\bigr],
\label{eq:sft} \\[4pt]
L_{\mathrm{CPO}} &= L_{\mathrm{pref}} + L_{\mathrm{SFT}}.
\label{eq:cpo}
\end{align}

Here:
\begin{itemize}
    \item $\sigma(\cdot)$ is the sigmoid function,
    \item $\beta$ is a positive scaling hyperparameter,
    \item $N$ is the total number of triplets,
    \item $y_i^{r} = y_i^{\mathrm{rejected}}$ (model output), $y_i^{c} = y_i^{\mathrm{chosen}}$ (human reference).
\end{itemize}

This objective enforces that $y_i^{c}$ should be ranked higher than $y_i^{r}$, while also maximizing the likelihood of $y_i^{c}$ under $\pi_{\theta}(\cdot)$. Further details can be found in \citet{xu_contrastive_2024}.


\subsection{Dataset}

For direct comparability with a prior study in MT domain adaptation (Anonymous, 2024), we borrow the exact same train, dev, and test set splits and divide our TM data into the same subsets of $\{1k, 2k, 5k, 10k, 14.7k\}$ segments.

\subsection{Training Configuration}

We adopt a batch size of 4 (with gradient accumulation = 8) and train for 1 epoch at a 1e-3 learning rate, saving checkpoints every 31 steps. This facilitates using checkpoints at 31, 62, 155, 310, and 459 steps to match the dataset sizes required for comparison. All runs use a single NVIDIA A100 GPU (80GB). Further implementation details appear in Appendix~\ref{app:impl-details}.

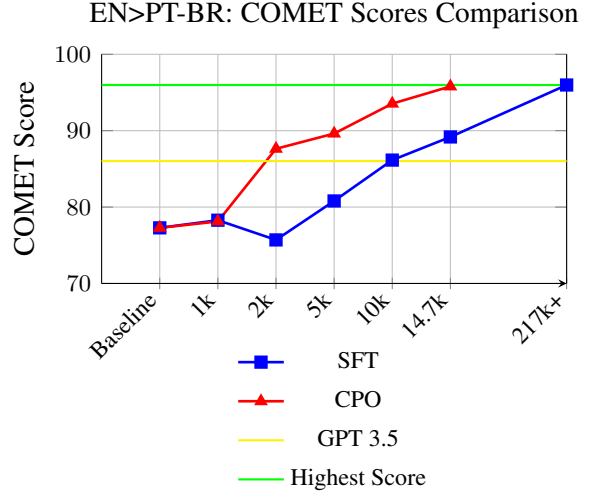
\begin{figure}[t]
    \centering
    \begin{tikzpicture}
        \begin{axis}[
            title={EN>PT-BR: COMET Scores Comparison},
            xlabel={Dataset Size},
            ylabel={COMET Score},
            xtick=data,
            xticklabels={Baseline, 1k, 2k, 5k, 10k, 14.7k,217k+}, 
            x tick label style={rotate=45, anchor=east, font=\footnotesize},
            y tick label style={font=\footnotesize},
            ymin=70, ymax=100,
            grid=major,
            width=\columnwidth,
            height=0.6\columnwidth,
            every axis plot/.append style={line width=1pt},
            axis x line=bottom,
            legend style={font=\footnotesize, at={(0.5,-0.25)}, anchor=north, draw=none, row sep=3pt}, 
        ]
            \addplot[
                color=blue,
                mark=square*,
            ]
            coordinates {
                (1, 77.28)
                (2, 78.28)
                (3, 75.70)
                (4, 80.80)
                (5, 86.15)
                (6, 89.18)
                (8, 95.98)
            };
            \addlegendentry{SFT}

            \addplot[
                color=red,
                mark=triangle*,
            ]
            coordinates {
                (1, 77.28)
                (2, 78.12)
                (3, 87.63)
                (4, 89.62)
                (5, 93.54)
                (6, 95.79)
            };
            \addlegendentry{CPO}

            \addplot[
                color=yellow,
                thick,
            ]
            coordinates {
                (0, 86.02) (8, 86.02)
            };
            \addlegendentry{GPT 3.5}

            \addplot[
                color=green,
                thick,
            ]
            coordinates {
                (0, 95.98) (8, 95.98)
            };
            \addlegendentry{Highest Score}
        \end{axis}
    \end{tikzpicture}
    \caption{Comparison of COMET Scores for EN>PTBR between SFT and CPO}
    \label{fig:ptbr_comet}
\end{figure}

3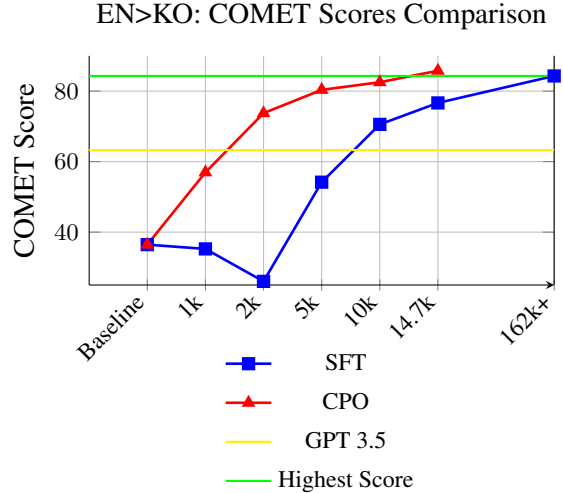
\begin{figure}[t]
    \centering
    \begin{tikzpicture}
        \begin{axis}[
            title={EN>KO: COMET Scores Comparison},
            xlabel={Dataset Size},
            ylabel={COMET Score},
            xtick=data,
            xticklabels={Baseline, 1k, 2k, 5k, 10k, 14.7k,162k+}, 
            x tick label style={rotate=45, anchor=east, font=\footnotesize},
            y tick label style={font=\footnotesize},
            ymin=25, ymax=90,
            grid=major,
            width=\columnwidth,
            height=0.6\columnwidth,
            every axis plot/.append style={line width=1pt},
            axis x line=bottom,
            legend style={font=\footnotesize, at={(0.5,-0.25)}, anchor=north, draw=none, row sep=3pt}, 
        ]
            \addplot[
                color=blue,
                mark=square*,
            ]
            coordinates {
                (1, 36.45)
                (2, 35.24)
                (3, 26.03)
                (4, 54.17)
                (5, 70.56)
                (6, 76.65)
                (8, 84.30)
            };
            \addlegendentry{SFT}

            \addplot[
                color=red,
                mark=triangle*,
            ]
            coordinates {
                (1, 36.45)
                (2, 56.95)
                (3, 73.74)
                (4, 80.35)
                (5, 82.52)
                (6, 85.78)
            };
            \addlegendentry{CPO}

            \addplot[
                color=yellow,
                thick,
            ]
            coordinates {
                (0, 63.28) (8, 63.28)
            };
            \addlegendentry{GPT 3.5}

            \addplot[
                color=green,
                thick,
            ]
            coordinates {
                (0, 84.30) (8, 84.30)
            };
            \addlegendentry{Highest Score}
        \end{axis}
    \end{tikzpicture}
    \caption{Comparison of COMET Scores for EN>KO between SFT and CPO}
    \label{fig:ko_comet}
\end{figure}

\subsection{Comparison with SFT}

Our methodology mirrors the SFT setup described in prior work (Anonymous, 2024), but replaces the standard cross-entropy objective with the CPO objective. This contrastive signal, we hypothesize, can achieve stronger domain alignment with fewer examples than SFT alone.


\paragraph{GPU Time and Sustainability.}
A prior experiment (Anonymous, 2024), trained on 14.7k segments with SFT, required 820 seconds on 4 NVIDIA A100 80GB GPUs (3,280 GPU-seconds total). By contrast, our CPO run on a single NVIDIA A100 80GB GPU required 1{,}598.90 GPU-seconds, a $\sim 51\%$ reduction in compute while achieving comparable or better effectiveness (see Table~\ref{tab:training_summary}). Using the same hardware and carbon model, our CPO run consumed 516.06 Wh and 173.39 gCO2e ($\approx 0.19$ tree-months). Assuming similar power draw and linear scaling, we estimate the SFT baseline at $\sim$1{,}058 Wh and $\sim$355 gCO2e for the 3{,}280 GPU-seconds. These figures indicate that, beyond being faster, the CPO recipe is also more sustainable in practice.

\section{Evaluation}

To evaluate model performance, we report BLEU \citep{papineni_bleu_2002}, chrF++ \citep{popovic_chrf_2017}, TER \citep{snover_study_2006}, and COMET \citep{rei_comet_2020}. We use the same metrics as prior work (Anonymous, 2024) for direct comparability. We select COMET as our primary metric for analysis given its strong correlation with human judgments \citep{zouhar-etal-2024-pitfalls}. Full results are in Appendix~\ref{app:results}.

As shown in Figure~\ref{fig:ptbr_comet} and Figure~\ref{fig:ko_comet}, our CPO-based approach provides substantial benefits for both PT-BR and KO. While larger training sets predictably lead to higher scores, our strategy of using synthetic preference pairs significantly narrows the data gap. Notably, with only 14.7k pairs, we achieve performance levels that approximate or exceed those obtained with more than 160k examples via SFT, demonstrating the data efficiency of this method.

\subsection{Discussion}

The findings underscore that injecting carefully crafted contrastive examples fosters domain-sensitive improvements without requiring massive data. By contrasting the model's own `rejected' outputs with `chosen' human-approved TM entries, our models learn to avoid undesirable constructs and internalizes stylistic and terminological nuances.

By generating on-policy negatives and combining ranking with likelihood in a single stage, our CPO recipe \emph{halves the GPU seconds} needed to reach strong in-domain quality and delivers \(\ge 6.61\) COMET gains over SFT when both are trained on the same 14.7k segments (see Figure~\ref{fig:ptbr_comet} and Figure~\ref{fig:ko_comet}). This constitutes a significant efficiency gain that shifts the cost curve for domain adaptation, which is particularly valued by industry.

\subsection{A Practical LSP Scenario}
Consider an organisation with 14.7k domain-specific sentences in a TM. Translators currently spend significant time post-editing raw MT outputs. Our CPO approach integrates the LLM’s \emph{initial} output as a “negative” example in each training instance, effectively capturing \emph{how} a translator would correct typical mistakes. Over time, the model learns to align new outputs with the style and terms in the curated TM, potentially reducing human PE overhead. With our proposed method, it is possible to achieve the MT performance that previously required more than 160k segments using SFT. Figure~\ref{fig:loop} illustrates this adaptive feedback loop.

\section{Conclusions}

This work investigated a data-efficient application of CPO to align LLMs to domain-specific standards for MT. Our approach harnesses the concept of ‘unlearning’: by contrasting a rejected translation (the model's own raw output) against a chosen one (a human reference from a TM), we teach the model to replace suboptimal constructs with domain-aligned alternatives. This process leverages preference feedback more directly than conventional SFT, allowing the model to internalize where and why certain translations fail.

Our experiments demonstrate that creating preference pairs from the model's own outputs against existing TMs can substantially reduce the amount of data required for domain adaptation. Using just 14.7k preference pairs, our models approach or surpass the performance of models trained with SFT on over 160k in-domain examples, highlighting the value of capturing rich preference information.

This CPO-based approach can be seamlessly integrated into future PEMT workflows, where newly post-edited translations yield a steady stream of preference pairs for continual training. This cycle promotes cost-effectiveness, allowing smaller, open-source models to be gradually shaped to exhibit near-custom quality without expensive, large-scale training from scratch. By incorporating contrastive signals, this method enhances model adaptability, making translation systems more robust and responsive to professional demands.

\section{Future Work}

\paragraph{Iterative CPO} Investigate a fully online learning pipeline whereby each new batch of post-edited examples is regularly integrated into the model training loop, leveraging the on-policy nature of our data generation.

\paragraph{Human Evaluation}While automatic metrics provide vital benchmarks, leveraging well-designed human assessments can capture subtleties in fluency, style, and domain accuracy that automated scores may miss.

\paragraph{Real Post-Edit Data} An under-explored resource is the vast repository of historical post-editing data held by LSPs. Beyond using TMs, integrating the triplet of “source text,” “pre-edited MT output,” and “human PEMT” provides a direct signal of the gap between automated translations and desired human quality. This gap captures the labour and expertise of human translators, serving as a powerful teaching signal. We plan to explore the use of this real, historical data for preference fine-tuning.

\section*{Limitations} 

While our CPO-based framework shows promise, it is subject to several limitations. First, our experiments are limited to two language pairs in a software localization context. This may hinder generalization to other domains (e.g., medical, legal) where stylistic norms and terminology differ dramatically.

Second, our method assumes TMs are a reliable, high-quality "gold standard". In practice, TMs may contain inconsistencies or outdated artifacts. While the goal of alignment still holds, performance may degrade if the reference TM suffers from significant noise.

\section*{Ethics Statement}
This research seeks to improve LLM outputs for domain-specific MT by leveraging existing high-quality TM data. We recognize that any language technology pipeline can introduce ethical concerns.

\paragraph{Translator Labour and Fairness.} Although our method aims to reduce the repetitive workload of PE, it also highlights potential shifts in translators’ roles. In principle, the approach should empower translators by reducing mundane edits and directing their expertise toward higher-level linguistic validation. However, any real-world deployment should include transparent communication about how translators’ post-edits will be used to refine MT systems, as well as fair compensation models that acknowledge the intellectual labour involved.


\bibliographystyle{style/acl_natbib}
\bibliography{mtsummit25}

\appendix
\newpage

\section{Implementation Details}
\label{app:impl-details}


\paragraph{Hardware and Environment.} We ran all experiments on a cluster equipped with:
\begin{itemize}
    \item 1 \textbf{NVIDIA A100} GPU (80GB) per run
    \item CUDA 12.5, PyTorch 2.0, Python 3.11.11
    \item \texttt{bitsandbytes} for 4-bit quantization
    \item \texttt{peft.LoraConfig} for LoRA injection
\end{itemize}

The code is based on \texttt{HuggingFace's CPOTrainer}, using a custom script 
(\texttt{run\_cpo.py}) that handles triplet parsing and training.

\vspace{5mm}
\begin{table}[h!]
\centering
\caption{Detailed Training Configuration}
\label{tab:trainconfig}
\resizebox{\linewidth}{!}{
\begin{tabular}{p{6cm} l}  
\toprule
\textbf{Parameter}             & \textbf{Value} \\
\midrule
\textit{Base Model Checkpoint} & \texttt{meta-llama/Meta-Llama-3-8B-Instruct} \\
\textit{Quantization}          & 4-bit, \texttt{bnb\_4bit\_quant\_type="nf4"} \\
\textit{Compute Dtype}         & \texttt{bfloat16} \\
\textit{LoRA Rank} ($r$)       & 64  \\
\textit{LoRA} $\alpha$         & 16  \\
\textit{LoRA Dropout}          & 0.1  \\
\textit{LoRA Target Modules}   & \texttt{\{q\_proj, v\_proj\}} \\
\textit{Training Epochs}       & 1 epoch \\
\textit{Batch Size}            & 4 (gradient accumulation = 8) \\
\textit{Learning Rate}         & 1e-3 \\
\textit{LR Scheduler Type}     & cosine \\
\textit{Seed}                  & 42 \\
\textit{Warmup Steps}          & 200 \\
\textit{Checkpoint Frequency}  & Every 31 steps \\
\textit{Logging Frequency}     & Every 100 steps \\
\bottomrule
\end{tabular}
}
\end{table}

We closely follow the hyperparameter scheme used in prior work (Anonymous, 2024) as much as possible, except we replace the standard cross-entropy loss with our CPO objective (see Section 3 in the main text) and we follow some hyperparameters from \citet{xu_contrastive_2024}.

\newpage
\section{Experimental Results}
\label{app:results}

\begin{table}[h]
    \centering
    \caption{SFT Results for PT-BR and KO}
    \label{tab:sft_results}
    \resizebox{\linewidth}{!}{ 
    \begin{tabular}{l l c c c c}
        \toprule
        \textbf{Language} & \textbf{Examples} & \textbf{BLEU} & \textbf{chrF++} & \textbf{TER} & \textbf{COMET} \\
        \midrule
        PT-BR & GPT 3.5 & 56.50 & 76.33 & 32.03 & 86.02 \\
        PT-BR & Baseline & 48.25 & 69.21 & 39.36 & 77.28 \\
        PT-BR & 1k & 48.00 & 69.34 & 40.11 & 78.28 \\
        PT-BR & 2k & 46.04 & 67.93 & 44.09 & 75.70 \\
        PT-BR & 5k & 49.73 & 69.92 & 38.03 & 80.80 \\
        PT-BR & 10k & 50.90 & 70.92 & 35.96 & 86.15 \\
        PT-BR & 14.7k & 53.42 & 73.07 & 32.92 & 89.18 \\
        PT-BR & 217k+ & 62.45 & 78.57 & 26.20 & 95.98 \\
        \midrule
        KO & GPT 3.5 & 33.07 & 49.72 & 60.60 & 63.28 \\
        KO & Baseline & 20.81 & 35.37 & 77.95 & 36.45 \\
        KO & 1k & 20.12 & 42.16 & 83.37 & 35.24 \\
        KO & 2k & 19.25 & 41.13 & 82.48 & 26.03 \\
        KO & 5k & 28.60 & 46.84 & 65.42 & 54.17 \\
        KO & 10k & 31.36 & 52.62 & 60.86 & 70.56 \\
        KO & 14.7k & 28.15 & 58.88 & 53.11 & 76.65 \\
        KO & 162k+ & 45.80 & 64.81 & 44.73 & 84.30 \\
        \bottomrule
    \end{tabular}
    }
\end{table}

\begin{table}[h]
    \centering
    \caption{CPO Results for PT-BR and KO}
    \label{tab:cpo_results}
    \resizebox{\linewidth}{!}{ 
    \begin{tabular}{l l c c c c}
        \toprule
        \textbf{Language} & \textbf{Examples} & \textbf{BLEU} & \textbf{chrF++} & \textbf{TER} & \textbf{COMET} \\
        \midrule
        PT-BR & 1k & 38.14 & 56.75 & 56.72 & 68.12 \\
        PT-BR & 2k & 52.25 & 72.32 & 34.23 & 87.63 \\
        PT-BR & 5k & 56.31 & 74.80 & 30.83 & 89.62 \\
        PT-BR & 10k & 60.74 & 77.43 & 28.30 & 93.54 \\
        PT-BR & 14.7k & 59.67 & 76.91 & 28.09 & 95.79 \\
        \midrule
        KO & 1k & 19.12 & 38.79 & 86.29 & 56.95 \\
        KO & 2k & 33.75 & 55.39 & 58.38 & 73.74 \\
        KO & 5k & 38.19 & 59.22 & 53.08 & 80.35 \\
        KO & 10k & 41.33 & 60.84 & 47.06 & 82.52 \\
        KO & 14.7k & 45.24 & 63.78 & 44.81 & 85.78 \\        
        \bottomrule
    \end{tabular}
    }
\end{table}

\section{Prompts and Special Tokens}
\label{sec:prompt}

\subsection{Special Token Descriptions}
\begin{quote}
\begin{footnotesize}
\begin{it}
$<|$begin\_of\_text$|>:$ This is equivalent to the BOS token.\\
$<|$eot\_id$|>:$ This signifies the end of the message in a turn.\\
$<|$start\_header\_id$|>$\{role\}$<|$end\_header\_id$|>:$ These tokens enclose the role for a particular message. The possible roles can be: system, user, assistant. \\
$<|$end\_of\_text$|>:$ This is equivalent to the EOS token.
\end{it}
\end{footnotesize}
\end{quote}

\subsection{Prompt}

\begin{quote}
\begin{footnotesize}
\begin{it}
$<|$begin\_of\_text$|>$
$<|$start\_header\_id$|>$system$<|$end\_header\_id$|>$\\ \\
You are a helpful AI assistant for translation from \{source\_language\} to \{target\_language\}. You MUST answer with the following JSON scheme: \{``translation'': ``string''\}
$<|$eot\_id$|>$\\
$<|$start\_header\_id$|>$user$<|$end\_header\_id$|>$ \\ \\
\{source\_sentence\}$<|$eot\_id$|>$$<|$start\_header\_id$|>$assistant$<|$end\_header\_id$|>$
\end{it}
\end{footnotesize}
\end{quote}

\subsection{Training Prompt}

\begin{quote}
\begin{footnotesize}
\begin{it}
$<|$begin\_of\_text$|>$$<|$start\_header\_id$|>$system$<|$end\_header\_id$|>$ \\

You are a helpful AI assistant for translation from \{source\_language\} to \{target\_language\}. You MUST answer with the following JSON scheme: \{``translation'': ``string''\} $<|$eot\_id$|>$ 

$<|$start\_header\_id$|>$user$<|$end\_header\_id$|>$
\{source\_sentence\}$<|$eot\_id$|>$ \\

$<|$start\_header\_id$|>$assistant$<|$end\_header\_id$|>$\textbf{\{target\_sentence\}$<|$end\_of\_text$|>$}
\end{it}
\end{footnotesize}
\end{quote}

\label{app:training-runs}
\begin{table*}[h!]
\centering
\caption{Summary of Training Runs}
\label{tab:training_summary}
\resizebox{2\columnwidth}{!}{%
\begin{tabular}{lcc}
\toprule
\textbf{Parameter} & \textbf{Training Run 1 (PT-BR)} & \textbf{Training Run 2 (KO)} \\
\midrule
\textit{Job ID} & 229296 & 229263 \\
\textit{Node} & g129 & g130 \\
\textit{GPU Model} & NVIDIA A100-SXM4-80GB & NVIDIA A100-SXM4-80GB \\
\textit{CUDA Version} & 12.5 & 12.5 \\
\textit{Batch Size (Train/Eval)} & 4 / 8 & 4 / 8 \\
\textit{Gradient Accumulation Steps} & 8 & 8 \\
\textit{Learning Rate} & 0.001 & 0.001 \\
\textit{LR Scheduler} & Cosine & Cosine \\
\textit{Quantization} & 4-bit BitsAndBytes & 4-bit BitsAndBytes \\
\textit{Epochs} & 1.0 & 1.0 \\
\textit{Dataset Size (Train/Validation)} & 14,700 / 0 & 14,700 / 0 \\
\textit{Final Train Loss} & 0.6351 & 0.9413 \\
\textit{Train Runtime (s)} & 1598.90 & 1560.08 \\
\textit{Samples per Second} & 9.194 & 9.423 \\
\textit{Steps per Second} & 0.287 & 0.294 \\
\textit{LoRA Adapter Path} & llama-aligned-cpo080225pt-br/checkpoint-xx & llama-aligned-cpo080225ko/checkpoint-xx \\
\textit{Translation Batch Size} & 5 & 5 \\
\textit{Max New Tokens} & 128 & 128 \\
\bottomrule
\end{tabular}%
}
\end{table*}

\begin{figure*}[h!]
    \centering
    \includegraphics[width=2\columnwidth]{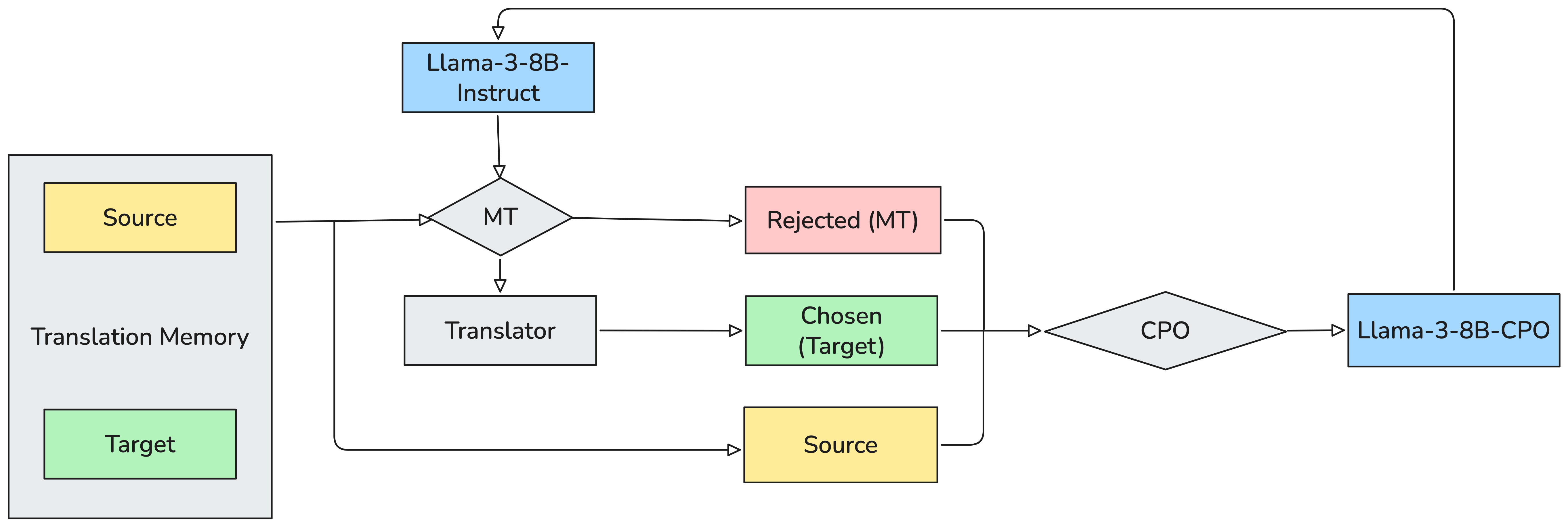}
    \caption{Adaptive MT: Symbiotic Human and LLM Translation Feedback Loop. Contrastive training on post-edited machine translation triplets.}
    \label{fig:loop}
\end{figure*}

\end{document}